\documentclass[10pt,twocolumn,letterpaper]{article}

\usepackage{cvpr}              

\usepackage{amsmath}
\usepackage{amssymb}
\usepackage{multirow}
\usepackage{booktabs}
\usepackage{graphicx}
\usepackage{url}
\usepackage[pagebackref,breaklinks,colorlinks,allcolors=cvprblue]{hyperref}
\usepackage[T1]{fontenc}
\usepackage{booktabs}
\usepackage{makecell}
\usepackage{array}
\usepackage{xcolor} 
\usepackage[accsupp]{axessibility}
\usepackage[capitalize]{cleveref}
\crefname{section}{Sec.}{Secs.}
\Crefname{section}{Section}{Sections}
\Crefname{table}{Table}{Tables}
\crefname{table}{Tab.}{Tabs.}

\definecolor{cvprblue}{rgb}{0.21,0.49,0.74}

\def\paperID{67} 
\def\confName{CVPR}
\def\confYear{2026}

\title{Scaling In-Context Segmentation with Hierarchical Supervision}

\author{Tidiane Camaret Ndir \qquad Marco Reisert \qquad Robin T. Schirrmeister\\
Division of Medical Physics, Department of Radiology,\\
Medical Center – University of Freiburg\\
Freiburg, Germany\\
{\tt\small \{tidiane.camaret.ndir, marco.reisert, robin.schirrmeister\}@uniklinik-freiburg.de}
}

\begin{document}
\maketitle

\begin{abstract}
In-context learning (ICL) enables medical image segmentation models to adapt to new anatomical structures from limited examples, reducing the clinical annotation burden. However, standard ICL methods typically rely on dense, global cross-attention, which scales poorly with image resolution. While recent approaches have introduced localized attention mechanisms, they often lack explicit supervision on the selection process, leading to redundant computation in non-informative regions. We propose PatchICL, a hierarchical framework that combines selective image patching with multi-level supervision. Our approach learns to actively identify and attend only to the most informative anatomical regions. Compared to UniverSeg, a strong global-attention baseline, PatchICL achieves competitive in-domain CT segmentation accuracy while reducing compute by 44\% at $512\times512$ resolution. On 35 out-of-domain datasets spanning diverse imaging modalities, PatchICL outperforms the baseline on 6 of 13 modality categories, with particular strength on modalities dominated by localized pathology such as OCT and dermoscopy. Training and evaluation code are available at \url{https://github.com/tidiane-camaret/ic_segmentation}.
\end{abstract}

\section{Introduction}
\label{sec:intro}

In-context learning (ICL) has emerged as a promising paradigm for medical image segmentation, allowing models to generalize to new anatomical structures from a handful of annotated examples without retraining. This capability is especially valuable in clinical settings, where acquiring dense annotations is expensive and time-consuming. However, deploying ICL methods on high-resolution medical volumes remains challenging due to the computational cost of dense cross-attention between target and context images.

Existing ICL approaches for segmentation fall into three broad categories, each with distinct limitations. Global methods such as UniverSeg~\cite{butoiUniverSegUniversalMedical2023} and Iris~\cite{gaoShowSegmentUniversal2025} employ dense cross-attention that scales quadratically with spatial resolution. Selective attention methods like Tyche~\cite{rakicTycheStochasticInContext2025} and EICSeg~\cite{xieEICSegUniversalMedical2026} improve focus but learn selection implicitly through segmentation loss alone. Patch-based approaches such as Medverse~\cite{huMedverseUniversalModel2025a} handle high resolutions via sliding windows but sever global context.

We propose PatchICL, a hierarchical framework that addresses these limitations through explicit patch selection with multi-level supervision. Our method processes images in a coarse-to-fine cascade, learning to identify and attend only to the most informative anatomical regions at each level. Unlike implicit attention methods, our patch selection is directly supervised, allowing early discarding of uninformative background before heavy processing.

Our contributions are as follows. First, we introduce a coarse-to-fine cascade with entropy-guided Gumbel-top-$K$ patch sampling that focuses computation on uncertain regions. Second, we apply multi-level supervision at each resolution stage, providing an explicit training signal for the selection process. Third, we demonstrate that PatchICL achieves competitive in-domain CT accuracy while reducing compute by 44\% at $512\times512$ resolution, and shows favorable scaling on out-of-domain high-resolution datasets.

\section{Related Work}
\label{sec:related}

\paragraph{Global in-context learning.}
Early adaptations of ICL for segmentation, such as UniverSeg~\cite{butoiUniverSegUniversalMedical2023}, utilize encoder--decoder architectures with dense cross-attention (CrossBlock) to fuse target and context features. Similarly, Iris~\cite{gaoShowSegmentUniversal2025} encodes context pairs into global task embeddings. These methods require processing the full resolution of both target and context images, resulting in quadratic complexity $\mathcal{O}((HW)^2)$ that prohibits scaling to large inputs.

\paragraph{Selective attention and retrieval.}
To address scalability, recent works have moved toward selective processing. Tyche~\cite{rakicTycheStochasticInContext2025} introduces stochasticity to explore multiple potential masks, effectively attending to different modes of the output distribution. EICSeg~\cite{xieEICSegUniversalMedical2026} employs retrieval mechanisms to align features before processing. While these methods focus attention on specific zones or features, selection is often implicit---learned solely through the final segmentation loss without direct supervision on what to select. Consequently, they often still require encoding the entire image volume before discarding irrelevant regions.

\paragraph{Patch-based and efficient approaches.}
Direct patch-based methods like Medverse~\cite{huMedverseUniversalModel2025a} use sliding window inference to handle high resolutions, while Patchwork~\cite{reisertDeepNeuralPatchworks2022a} employs a hierarchical stacking strategy to iteratively sample uncertain regions. Pure sliding windows treat patches independently, severing global context.

\paragraph{Positioning.}
Unlike implicit attention methods, we employ explicit patch selection trained via multi-level supervision. This allows our model to discard uninformative background regions before heavy processing, retaining global context efficiency without the computational cost of full-volume attention.

\section{Method}
\label{sec:method}

\noindent \textbf{Input:} Target image $I^t \in \mathbb{R}^{H \times W}$ and context set $\{(I^c_i, L^c_i)\}_{i=1}^{N_c}$ of image--label mask pairs.

\noindent \textbf{Output:} Predicted segmentation mask $\hat{L}^t$ at original resolution.

\subsection{Overview}
\label{sec:overview}

Our method processes images through a coarse-to-fine cascade of $M$ resolution levels $\{r_1, \ldots, r_M\}$ (\eg, $24 \rightarrow 96 \rightarrow 128$). At each level: (1)~\textit{Sampling}: select $K$ patches via Gumbel-top-$K$ sampling, using uncertainty from the previous level as sampling weights; (2)~\textit{Attention}: process target and context patches jointly through transformer layers; (3)~\textit{Aggregation}: combine potentially overlapping patch predictions via averaging; and (4)~\textit{Refinement}: additively fuse with upsampled previous-level predictions.

\subsection{Patch Sampling}
\label{sec:sampling}

At level $\ell > 1$, target patch sampling is guided by the entropy of the previous level's prediction:
\begin{align}
    w_\ell(x, y) &= H(p_{\ell-1}(x,y)), \nonumber \\
    H(p) &= -p\log p - (1-p)\log(1-p), \label{eq:entropy}
\end{align}
where $p_{\ell-1} = \sigma(\text{logits}_{\ell-1})$, focusing computation on uncertain regions. Patch-level sampling weights are obtained by averaging entropy over each candidate patch region. These weights are passed through a Gumbel-top-$K$ sampler~\cite{kool2019stochasticbeamsthemgumbeltopk}:
\begin{equation}
    S_\ell = \operatorname{top\text{-}K}\bigl(\log \bar{w}_\ell(k) + g_k,\; K\bigr), \quad g_k \sim \text{Gumbel}(0,1),
    \label{eq:gumbel}
\end{equation}
where $\bar{w}_\ell(k)$ denotes the mean entropy weight for candidate patch $k$ and $g_k$ is an independent Gumbel noise sample. The temperature of the Gumbel distribution is fixed at $\tau = 1$ throughout training.

For context patches, sampling weights are derived from the boundary proximity of the ground-truth mask: we compute a distance transform of the label and assign higher weight to patches whose mean distance to the foreground boundary is small. This encourages context patches to cover the most structurally informative regions.

\subsection{Feature Extraction and Encoding}
\label{sec:features}

Input images are encoded using the frozen encoder from UniverSeg~\cite{butoiUniverSegUniversalMedical2023}. A lightweight CNN encoder projects these features to embeddings $\mathbf{z}_k \in \mathbb{R}^{d}$, producing skip connections for the decoder. To enable a single backbone across all resolution levels, we condition on resolution via sinusoidal encoding, allowing generalization to unseen resolutions.

\subsection{Patch-level Attention}
\label{sec:attention}

Target and context patch embeddings are processed jointly through $L$ transformer layers with: (1)~2D rotary position embeddings (RoPE) encoding spatial patch coordinates $(y, x)$, enabling reasoning about relative patch positions; (2)~per-layer learnable type embeddings distinguishing target from context patches; and (3)~bidirectional attention, where all patches attend to all others. A CNN decoder reconstructs patch-level predictions from the attended features.

\subsection{Aggregation and Level Combination}
\label{sec:aggregation}

Patch predictions are aggregated using simple averaging over overlapping regions:
\begin{equation}
    \hat{L}^t_\ell(x,y) = \frac{\sum_{k} \mathbf{1}_{(x,y) \in P_k} \cdot p_k(x,y)}{\sum_{k} \mathbf{1}_{(x,y) \in P_k}},
    \label{eq:aggregation}
\end{equation}
where $\mathbf{1}_{(x,y) \in P_k}$ indicates pixel $(x,y)$ is covered by patch $k$.

Level predictions are combined via additive fusion within the coverage mask:
\begin{equation}
    \hat{L}^t_{\text{comb}} = \hat{L}^t_{\ell-1,\uparrow} + M_\ell \odot \hat{L}^t_\ell,
    \label{eq:fusion}
\end{equation}
where $\hat{L}^t_{\ell-1,\uparrow}$ is the upsampled previous prediction and $M_\ell$ is the binary coverage mask indicating which pixels were sampled at level $\ell$.

\begin{figure}[t]
\centering
\includegraphics[width=\linewidth]{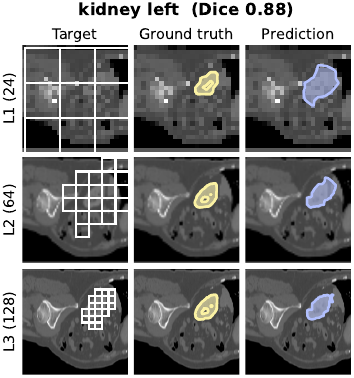}
\caption{%
  Coarse-to-fine patch selection across resolution levels on subject~0013 of the TotalSegmentator CT dataset.
  Columns show the target image with patch boxes, ground-truth overlay,
  and prediction overlay.
  Rows correspond to cascade levels at increasing resolution.%
  }
\label{fig:patch_selection_qualitative}
\end{figure}

\subsection{Training}
\label{sec:training}

The model is trained with supervision at each level. Each level loss combines binary cross-entropy and soft Dice:
\begin{equation}
    \mathcal{L}_\text{combined}^\ell = \mathcal{L}_\text{BCE}^\ell + \mathcal{L}_\text{Dice}^\ell,
    \label{eq:level_loss}
\end{equation}
applied to the combined prediction $\hat{L}^t_{\text{comb},\ell}$ against the ground-truth mask at the corresponding resolution. The total loss sums over all levels with uniform weights:
\begin{equation}
    \mathcal{L} = \sum_{\ell=1}^{M} \mathcal{L}_\text{combined}^\ell.
    \label{eq:loss}
\end{equation}

\begin{figure}[t]
\centering
\includegraphics[width=\linewidth]{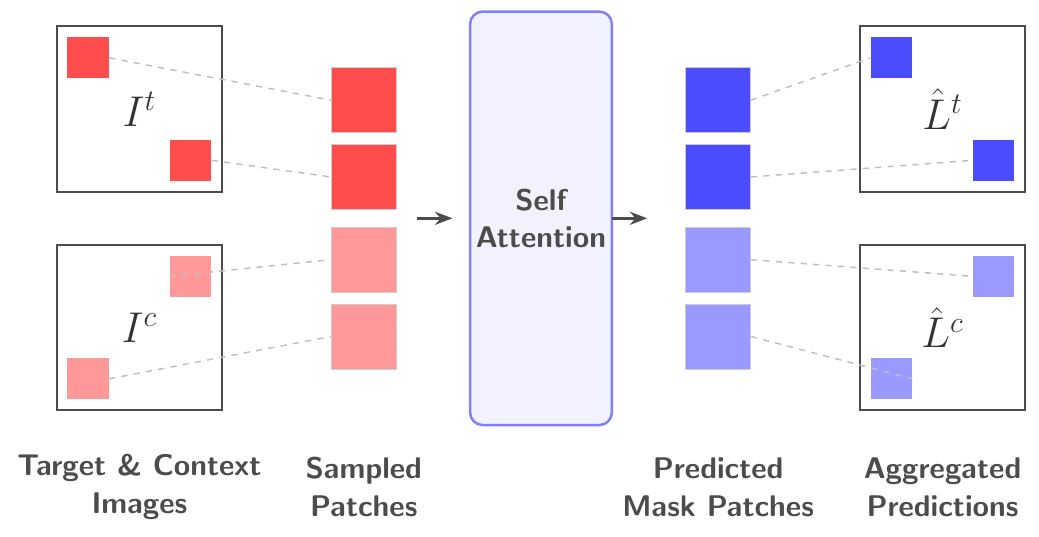}
\caption{Overview of the PatchICL architecture. Image patches are selectively sampled on the target and context images, and mask predictions are aggregated.}
\label{fig:architecture}
\end{figure}

\begin{figure}[t]
\centering
\includegraphics[width=\linewidth]{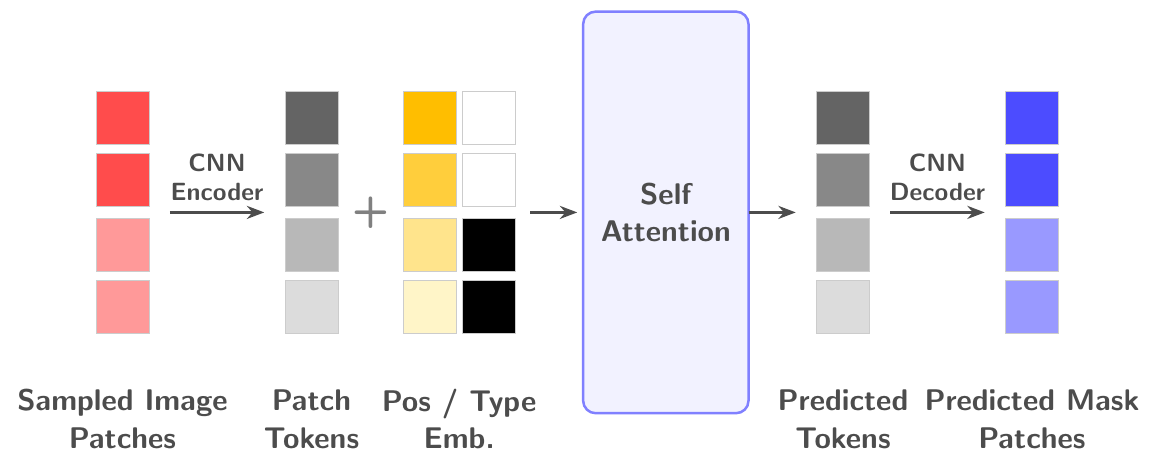}
\caption{Patch-level attention mechanism. Target and context patches are processed jointly through transformer layers with 2D RoPE, type embeddings, and bidirectional attention.}
\label{fig:patch_attn}
\end{figure}

\section{Experiments}
\label{sec:experiments}

\subsection{Dataset and Preprocessing}
\label{sec:dataset}

We train PatchICL on TotalSegmentator~\cite{wasserthalTotalSegmentatorRobustSegmentation2023}, a large-scale CT dataset with various annotated anatomical structures. To test in-context generalization, we split anatomical classes into 47 structures for training and 45 held-out structures for validation. This ensures the model must generalize to entirely unseen anatomical categories at test time, not just unseen patients. \cref{tab:category_definitions} lists all structures grouped by reporting category.

\paragraph{2D slice extraction.}
We extract 2D slices from 3D volumes along all three anatomical axes (axial, coronal, sagittal), increasing data diversity and providing multi-view perspectives of each structure. Slices are filtered by coverage: only slices with $\geq 30$ foreground pixels are retained, preventing training on uninformative near-empty slices.

\paragraph{Context selection.}
Context slices are preferentially drawn from the same case as the target, providing anatomically consistent support. When insufficient same-case examples exist, we fall back to other patients with the same structure. We use $N_c = 3$ context pairs during training and validation.

\paragraph{Evaluation protocol.}
We assess our method across three distinct generalization settings. For \textit{in-domain} performance, we evaluate on TotalSegmentator CT using the 45 held-out anatomical classes on validation cases. To test \textit{cross-modality} transfer, we utilize the TotalSegmentator MRI dataset~\cite{dantonoliTotalSegmentatorMRIRobust2025} (all 46 classes, validation split). Finally, we measure \textit{out-of-domain} generalization on MedSegBench~\cite{kusMedSegBenchComprehensiveBenchmark2024}, challenging the model with 35 diverse datasets spanning modalities such as ultrasound, X-ray, and microscopy.

\begin{table}[t]
\centering
\caption{Anatomical structure categories for TotalSegmentator
evaluation. Left/right pairs abbreviated (L/R).
Unmarked structures appear in both CT and MRI;
\textsuperscript{c}\,=\,CT only,
\textsuperscript{m}\,=\,MRI only.}
\label{tab:category_definitions}
\small
\setlength{\tabcolsep}{4pt}
\begin{tabular}{@{}p{1.8cm} p{5.4cm}@{}}
\toprule
\textbf{Category} & \textbf{Structures} \\
\midrule
Organs (Abd./ Pelvis)
& adrenal gland (L/R), colon, duodenum, esophagus,
  gallbladder, kidney (L/R),
  kidney cyst\textsuperscript{c} (L/R),
  liver, pancreas, prostate, small bowel, spleen,
  stomach, urinary bladder \\
\addlinespace[3pt]
Organs (Thrx./ Head/Spine)
& brain, heart, spinal cord,
  atrial appendage\textsuperscript{c} (L),
  lung lobes\textsuperscript{c} (5),
  lung\textsuperscript{m} (L/R),
  thyroid gland\textsuperscript{c},
  trachea\textsuperscript{c} \\
\addlinespace[3pt]
Bones (Ribs/ Sternum)
& costal cartilages\textsuperscript{c},
  ribs\textsuperscript{c} L\,1--12,
  ribs\textsuperscript{c} R\,1--12,
  sternum\textsuperscript{c} \\
\addlinespace[3pt]
Bones (Limbs/ Shld./Pelvis)
& femur (L/R), hip (L/R), humerus (L/R),
  clavicula\textsuperscript{c} (L/R),
  scapula\textsuperscript{c} (L/R),
  skull\textsuperscript{c},
  fibula\textsuperscript{m},
  tibia\textsuperscript{m} \\
\addlinespace[3pt]
Muscles (Trunk)
& autochthon (L/R),
  gluteus max./med./min.\ (L/R),
  iliopsoas (L/R) \\
\addlinespace[3pt]
Muscles (Thigh)
& quadriceps femoris\textsuperscript{m} (L/R),
  sartorius\textsuperscript{m} (L/R),
  thigh med./post.\ comp.\textsuperscript{m} (L/R) \\
\addlinespace[3pt]
Vessels
& aorta, iliac artery (L/R), iliac vena (L/R),
  inferior vena cava, portal/splenic vein,
  brachiocephalic trunk\textsuperscript{c},
  brachiocephalic vein\textsuperscript{c} (L/R),
  common carotid art.\textsuperscript{c} (L/R),
  pulmonary vein\textsuperscript{c},
  subclavian art.\textsuperscript{c} (L/R),
  superior vena cava\textsuperscript{c} \\
\bottomrule
\end{tabular}
\end{table}
\subsection{Scaling with Input Resolution}
\label{sec:scaling}
A key motivation for PatchICL is efficient scaling to high-resolution inputs.
We therefore begin by examining how accuracy and computational cost evolve
with input resolution on the TotalSegmentator CT validation set.
 
\begin{figure}[t]
  \centering
  \includegraphics[width=\linewidth]{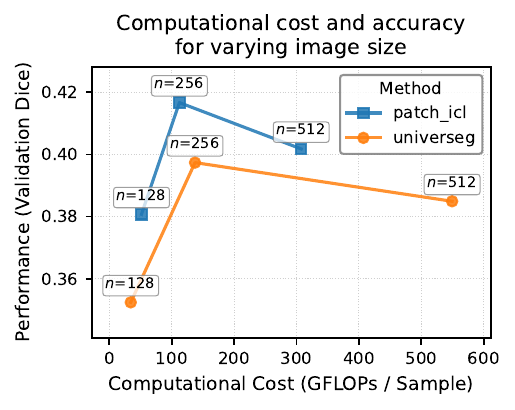}
  \caption{Dice score and computational cost (GFLOPs) as a function of
    input image resolution on TotalSegmentator CT\@.
    PatchICL achieves competitive accuracy with significantly reduced
    compute at higher resolutions.}
  \label{fig:dice_flops_per_input_size}
\end{figure}
 
\cref{fig:dice_flops_per_input_size} shows that PatchICL achieves higher
Dice scores at all tested resolutions while requiring significantly less
compute at high resolutions.
At $512\times512$, PatchICL uses 308~GFLOPs compared to UniverSeg's
549~GFLOPs---a 44\% reduction---while maintaining a Dice advantage (see
figure for exact values).
Both methods peak at $256\times256$ resolution, suggesting diminishing
returns from higher resolutions on this benchmark.
The favorable scaling of PatchICL stems from its patch-based attention,
which processes only selected regions rather than the full image.
This efficiency advantage motivates the detailed accuracy analysis that
follows.
\subsection{In-Domain CT Segmentation}
\label{sec:ct_results}

Having established PatchICL's computational advantage, we now examine per-category accuracy on the in-domain TotalSegmentator CT benchmark.

\begin{table}[t]
\centering
\caption{Dice scores on TotalSegmentator CT validation cases and held-out classes (mean $\pm$ std). $N$ denotes the number of evaluated slices. Best per category in \textbf{bold}.}
\label{tab:dice_totalseg_ct}
\small
\setlength{\tabcolsep}{4pt} 
\begin{tabular}{@{}l r c c@{}} 
\toprule
Category & $N$ & PatchICL & UniverSeg \\
\midrule
\makecell[l]{Bones \\ (Limbs/Shldr./Pelvis)} & 1,992 & \textbf{0.572 $\pm$ 0.296} & 0.523 $\pm$ 0.331 \\ \addlinespace
\makecell[l]{Bones \\ (Spine)}               & 780   & 0.464 $\pm$ 0.271 & \textbf{0.552 $\pm$ 0.281} \\ \addlinespace
\makecell[l]{Organs \\ (Thorax/Head/Spine)}  & 1,077 & 0.568 $\pm$ 0.328 & \textbf{0.584 $\pm$ 0.316} \\ \addlinespace
\makecell[l]{Organs \\ (Abd./Pelvis)}        & 8,498 & 0.461 $\pm$ 0.314 & \textbf{0.470 $\pm$ 0.335} \\ \addlinespace
Muscles                                      & 4,494 & 0.444 $\pm$ 0.328 & \textbf{0.454 $\pm$ 0.332} \\ \addlinespace
\makecell[l]{Bones \\ (Ribs/Sternum)}        & 13,620 & \textbf{0.390 $\pm$ 0.308} & 0.321 $\pm$ 0.334 \\ \addlinespace
Vessels                                      & 8,298 & 0.340 $\pm$ 0.326 & \textbf{0.348 $\pm$ 0.339} \\
\midrule
\textbf{Overall}                             & 38,759 & \textbf{0.417 $\pm$ 0.321} & 0.397 $\pm$ 0.343 \\
\bottomrule
\end{tabular}
\end{table}

\cref{tab:dice_totalseg_ct} shows that PatchICL outperforms UniverSeg overall (0.417 vs.\ 0.397), winning 2 of 7 categories and achieving a higher score on 27 of 45 individual classes (60\%). UniverSeg leads on the remaining 5 categories, with its largest advantage on Bones (Spine).

\begin{figure}[t]
\centering
\includegraphics[width=\linewidth]{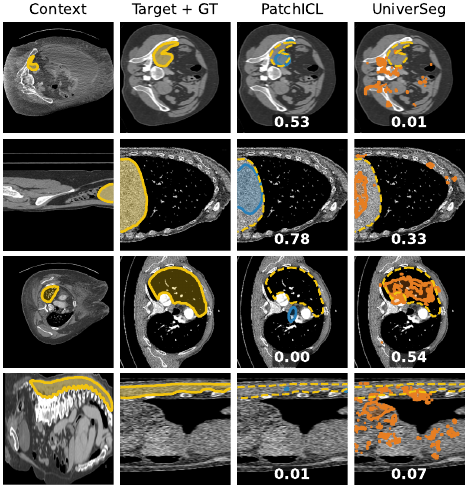}
\caption{%
    Qualitative comparison on TotalSegmentator CT held-out classes.
  Columns: context pair, target with \textcolor{yellow!80!black}{\textbf{yellow}} ground truth,
  \textcolor{blue!70!black}{\textbf{blue}} PatchICL prediction,
  \textcolor{orange!80!black}{\textbf{orange}} UniverSeg prediction.
  Dashed yellow contours show GT on prediction columns. Dice scores inset.
  Top rows: PatchICL wins (iliopsoas, liver); bottom rows: UniverSeg wins (lung lobe, autochthon).%
  }
\label{fig:qualitative_comparison}
\end{figure}

\begin{figure*}[t]
\centering
\begin{minipage}[t]{0.4\textwidth}
  \centering
  \includegraphics[width=\linewidth]{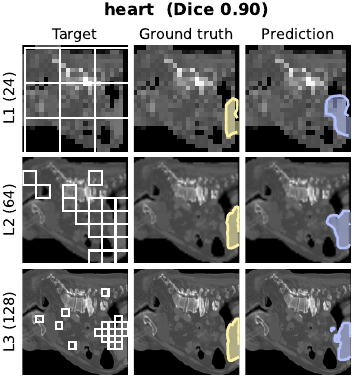}
  \subcaption{Case 0236 (heart)}
\end{minipage}%
\hfill
\begin{minipage}[t]{0.4\textwidth}
  \centering
  \includegraphics[width=\linewidth]{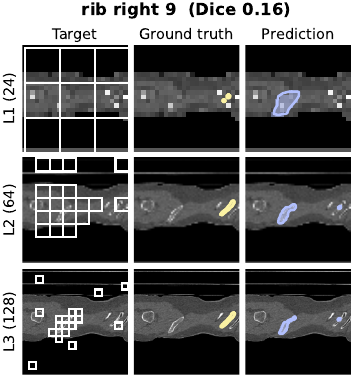}
  \subcaption{Case 0324 (Ribs right 9)}
\end{minipage}%
\caption{%
  Additional coarse-to-fine patch selection examples on TotalSegmentator CT.
  Columns show the target image with patch boxes, ground-truth overlay,
  and prediction overlay.
  Rows correspond to cascade levels at increasing resolution.
}
\label{fig:supp_patch_selection}
\end{figure*}

 \cref{fig:supp_patch_selection} provides additional examples of the
coarse-to-fine patch selection cascade, complementing
\cref{fig:patch_selection_qualitative}.

\subsection{Cross-Modality Generalization (MRI)}
\label{sec:mri_results}

We next evaluate whether the accuracy--efficiency trade-off observed on CT transfers to an unseen imaging modality. We test on TotalSegmentator MRI~\cite{dantonoliTotalSegmentatorMRIRobust2025}, which contains 46 anatomical structures across abdominal and musculoskeletal regions.

\begin{table}[t]
\centering
\caption{Dice scores on TotalSegmentator MRI (mean $\pm$ std). $N$ denotes the number of evaluated slices. The overall row averages across all individual classes; category rows are shown for interpretability but contain varying numbers of classes. Best per category in \textbf{bold}.}
\label{tab:dice_totalseg_mri}
\small
\setlength{\tabcolsep}{3pt}
\begin{tabular}{@{}l r c c@{}}
\toprule
Category & $N$ & PatchICL & UniverSeg \\
\midrule
\makecell[l]{Organs \\ (Thorax/Head/Spine)}  & \phantom{0}936 & \textbf{0.550 $\pm$ 0.291} & 0.547 $\pm$ 0.294 \\ \addlinespace
\makecell[l]{Muscles \\ (Trunk)}             & 1\,396 & 0.447 $\pm$ 0.304 & \textbf{0.488 $\pm$ 0.305} \\ \addlinespace
\makecell[l]{Organs \\ (Abd./Pelvis)}        & 2\,072 & 0.436 $\pm$ 0.298 & \textbf{0.459 $\pm$ 0.307} \\ \addlinespace
\makecell[l]{Bones \\ (Spine)}               & \phantom{0}441 & 0.405 $\pm$ 0.234 & \textbf{0.500 $\pm$ 0.236} \\ \addlinespace
\makecell[l]{Bones \\ (Limbs/Pelvis)}        & \phantom{0}520 & 0.417 $\pm$ 0.340 & \textbf{0.423 $\pm$ 0.350} \\ \addlinespace
Vessels                                      & \phantom{0}831 & 0.381 $\pm$ 0.312 & \textbf{0.398 $\pm$ 0.326} \\
\midrule
\textbf{Overall}                             & 6\,196 & 0.444 $\pm$ 0.304 & \textbf{0.471 $\pm$ 0.310} \\
\bottomrule
\end{tabular}
\end{table}

As shown in \cref{tab:dice_totalseg_mri}, UniverSeg outperforms PatchICL on overall Dice by +2.9\%. The largest category-level gaps appear on Bones (Spine) ($-$8.2\%) and Muscles (Trunk) ($-$4.1\%), consistent with CT findings that UniverSeg excels on complex articulated anatomy.

\begin{figure}[t]
\centering
\includegraphics[width=\linewidth]{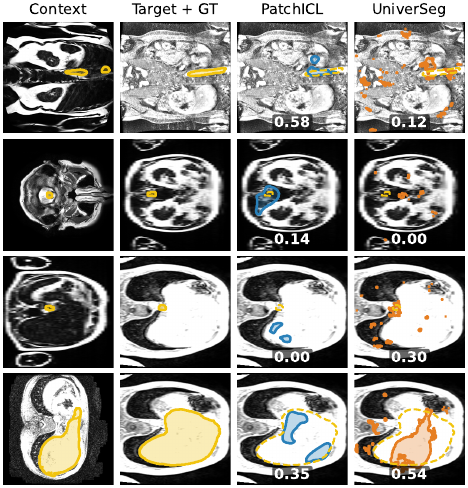}
\caption{%
  Cross-modality comparison on TotalSegmentator MRI (both methods trained on CT only).
  Columns: context pair, target with \textcolor{yellow!80!black}{\textbf{yellow}} ground truth,
  \textcolor{blue!70!black}{\textbf{blue}} PatchICL prediction,
  \textcolor{orange!80!black}{\textbf{orange}} UniverSeg prediction.
  Dice scores inset.%
}
\label{fig:qualitative_comparison_mri}
\end{figure}
\subsection{Out-of-Domain Generalization (MedSegBench)}
\label{sec:ood_results}

Finally, we evaluate generalization beyond CT imaging on MedSegBench~\cite{kusMedSegBenchComprehensiveBenchmark2024}, a comprehensive benchmark spanning 35 datasets across diverse modalities including ultrasound, X-ray, endoscopy, dermoscopy, fundus imaging, and microscopy. Both methods were trained exclusively on TotalSegmentator CT.

\begin{table}[t]
\centering
\caption{Dice scores on MedSegBench by imaging modality (mean $\pm$ std), using 3 context pairs. $N$ denotes the number of evaluated samples. Best per modality in \textbf{bold}.}
\label{tab:medsegbench_category}
\small
\setlength{\tabcolsep}{3pt}
\begin{tabular}{@{}p{2.4cm}rcc@{}}
\toprule
Modality & $N$ & PatchICL & UniverSeg \\
\midrule
X-Ray           &     \phantom{0,0}11 & 0.727 $\pm$ 0.045 & \textbf{0.896 $\pm$ 0.025} \\
Dermoscopy      &  \phantom{0,}209 & \textbf{0.643 $\pm$ 0.204} & 0.578 $\pm$ 0.190 \\
Chest X-Ray     &          2\,581 & 0.610 $\pm$ 0.174 & \textbf{0.684 $\pm$ 0.216} \\
Microscopy      &  \phantom{0,}263 & \textbf{0.539 $\pm$ 0.252} & 0.495 $\pm$ 0.285 \\
Ultrasound      &          1\,209 & 0.484 $\pm$ 0.241 & \textbf{0.553 $\pm$ 0.261} \\
MRI             &  \phantom{0,}147 & 0.433 $\pm$ 0.270 & \textbf{0.565 $\pm$ 0.211} \\
Nuclei          &   \phantom{0,0}67 & 0.344 $\pm$ 0.150 & \textbf{0.356 $\pm$ 0.181} \\
Fundus          &   \phantom{0,0}26 & \textbf{0.279 $\pm$ 0.212} & 0.223 $\pm$ 0.185 \\
OCT             &  \phantom{0,}101 & \textbf{0.272 $\pm$ 0.223} & 0.140 $\pm$ 0.188 \\
Nuclear Cell    &          1\,757 & 0.254 $\pm$ 0.120 & \textbf{0.418 $\pm$ 0.179} \\
Endoscopy       &  \phantom{0,}697 & \textbf{0.174 $\pm$ 0.231} & 0.147 $\pm$ 0.220 \\
CT              &  \phantom{0,}272 & \textbf{0.121 $\pm$ 0.178} & 0.098 $\pm$ 0.165 \\
Pathology       &   \phantom{0,0}40 & 0.118 $\pm$ 0.146 & \textbf{0.129 $\pm$ 0.146} \\
\midrule
\textbf{Overall} &         7\,380 & 0.430 $\pm$ 0.260 & \textbf{0.500 $\pm$ 0.286} \\
\bottomrule
\end{tabular}
\end{table}

\cref{tab:medsegbench_category} stratifies MedSegBench results by modality category. PatchICL outperforms UniverSeg on 6 of 13 modalities, with the largest gains on OCT (+13.2\%) and dermoscopy (+6.6\%), where localized features and fine-grained textures dominate. UniverSeg leads on the remaining 7 modalities, with its strongest advantages on X-Ray ($-$16.8\%) and Nuclear Cell ($-$16.4\%), both characterized by high-contrast, globally structured targets. Overall, PatchICL tends to excel on modalities with subtle, spatially localized pathology, while UniverSeg benefits from modalities with well-defined, high-contrast structures.

\section{Conclusion}
\label{sec:conclusion}

We presented PatchICL, a scalable hierarchical framework for in-context medical image segmentation. By explicitly operating at multiple resolution levels with entropy-guided patch selection, our approach reduces computational overhead while preserving global context. Our evaluations demonstrate that PatchICL achieves competitive performance on in-domain CT and cross-modality MRI tasks, excelling particularly on repetitive skeletal structures and localized out-of-domain pathologies. While the UniverSeg baseline maintains a slight advantage on complex articulated anatomies, PatchICL offers a favorable accuracy--compute trade-off, reducing compute by 44\% at $512\times512$ resolution without sacrificing performance. Future work will explore learned patch sampling strategies to better exploit specific medical structures and extend the cascade with adaptive level selection.

\section*{Acknowledgments}
This work was funded by the Deutsche Forschungsgemeinschaft (DFG, German Research Foundation) – Projektnummer(n) * / SPP 2177.

{
    \small
    \bibliographystyle{ieeenat_fullname}
    \bibliography{main}
}

\end{document}